\documentclass{article}
\usepackage[final]{colm2026_conference}

\usepackage{microtype}
\usepackage{hyperref}
\usepackage{url}
\usepackage{booktabs}
\usepackage{longtable}
\usepackage{subcaption}

\usepackage{hyperref}

\usepackage{amsmath}
\usepackage{amssymb}
\usepackage{mathtools}
\usepackage{amsthm}

\usepackage[capitalize,noabbrev]{cleveref}

\theoremstyle{plain}

\theoremstyle{definition}

\theoremstyle{remark}

\usepackage[textsize=tiny]{todonotes}

\usepackage{lineno}

\definecolor{darkblue}{rgb}{0, 0, 0.5}
\hypersetup{colorlinks=true, citecolor=darkblue, linkcolor=darkblue, urlcolor=darkblue}

\title{Creativity and Hallucinations Share Mechanisms in LLMs}

\author{
\\
\parbox[t]{0.23\textwidth}{\centering
\textbf{Neta Glazer}\\[0.2em]
{\small Bar-Ilan University}
}
\hfill
\parbox[t]{0.23\textwidth}{\centering
\textbf{Hadas Orgad}\\[0.2em]
{\small Kempner Institute}\\
{\small Harvard University}
}
\hfill
\parbox[t]{0.23\textwidth}{\centering
\textbf{Lenny Aharon}\\[0.2em]
{\small Columbia University}
}
\hfill
\parbox[t]{0.23\textwidth}{\centering
\textbf{Ethan Fetaya}\\[0.2em]
{\small Bar-Ilan University}
}
\\[1.7em]
\\
\texttt{neta.glazer@biu.ac.il}
}

\begin{document}

\ifcolmsubmission
\linenumbers
\fi

\maketitle

\begin{abstract}
A potential connection between creativity and hallucinations in large language models has been previously discussed, but their relationship is still not well understood and current evidence is primarily correlational. In this work, we study this question using mechanistic interpretability and targeted internal interventions. We localize components associated with creativity by predicting creativity metrics across established creativity benchmarks. We then intervene on these components and measure the resulting changes in factuality benchmarks. Across settings, we observe a component-level trade-off: selectively attenuating a small subset of creativity-linked components reduces creativity metrics while improving factual performance under fixed decoding. These results provide intervention-based evidence that internal computations supporting creative generation can also contribute to hallucination-prone behavior. We also observe strong alignment between creativity steering directions and hallucination-associated directions across models, providing complementary geometric evidence for a shared representational axis underlying this trade-off.
Together, these findings suggest that the creativity–hallucination trade-off is not merely a behavioral tendency but a structural property of how these models internally represent and generate text.
\end{abstract}

\section{Introduction}

Large language models (LLMs) are increasingly used for a wide variety of applications from ideation and story writing, to coding and mathematical reasoning \citep{brown2020language,ouyang2022training,achiam2023gpt, spangher2025creative}.
At the same time, these models can sometimes produce fluent but factually ungrounded content, often referred to as hallucinations or confabulations \citep{maynez2020faithfulness,ji2023survey,huang2025survey}.
Prior work has suggested a possible connection between creativity and hallucination in LLMs \citep{chakrabarty2024art,franceschelli2025creativity}, but this relationship remains poorly understood and is largely characterized by correlational observations at the level of model outputs.
If such a connection exists at the level of internal mechanisms, it would imply that the capacity for creative generation is structurally entangled with hallucination-prone behavior in current architectures, making the trade-off a fundamental property of how these models are built, rather than an artifact of how they are prompted or evaluated.

Creativity in LLMs has been defined in many different ways across the literature \citep{franceschelli2025creativity, hou2025creativityprism, lu2025benchmarking}.
In this work, we focus specifically on creative writing: the generation of novel, open-ended narratives.
This is a natural setting for studying the creativity–hallucination connection because producing ungrounded content is not only common in creative writing but often desirable. Consider, for example, the Harry Potter books: the appeal of these narratives depends strongly on a combination of facts and fantastical elements, and it is precisely these fabricated details that are central to the stories' appeal and what a successful LLM should try to mimic in this case. 

In these settings, producing ungrounded content is not a failure but the goal. It stands to reason that this type of creativity can leak to applications where it is not desirable, manifesting as hallucination.

Recent work has begun to examine this connection empirically, showing that creativity and hallucination move together across decoding settings \citep{he2025shakespearean_sparks}, that hallucination-reduction methods affect creativity non-uniformly \citep{banerjee2025less_hallucination_less_creativity}, and that the relationship is nonlinear and task-dependent \citep{yang2025hicbench}.
These results establish that the two behaviors co-vary, but they are measured entirely at the level of model outputs, leaving two questions open: whether the relationship is causal, and how it is reflected in the model's internal representations.

To close this gap, we study the relationship between creativity and factuality directly within the model's activation space.
We identify model components critical to creative writing, and show that \textbf{ablating components that contribute to creativity improves factuality}, providing intervention-based evidence that creativity and hallucinations rely on shared internal mechanisms.

We then observe this connection geometrically:
\textbf{steering vectors that enhance creativity reduce factuality and vice versa}, and the directions associated with creativity and hallucination are closely aligned in representation space. 
Across multiple experiments, benchmarks, and models, we find consistent evidence for a component- and representation-level trade-off between these two capabilities.
Together, these results suggest that the creativity–factuality tension is not merely visible in model outputs, but is rooted in the model's mechanisms.


Understanding this relationship has practical implications.
Our findings suggest that hallucinations cannot be treated as an isolated failure mode: any intervention targeting creativity-linked components will, at least partially, come at the cost of factuality.
This matters not only for understanding the limits of current models, but for motivating future designs that more explicitly account for the interplay between creative and factual generation.
Together, our findings reframe the creativity–hallucination trade-off as a structural signature of how current language models internally represent and generate text.


\begin{figure}
    \centering
    \includegraphics[width=\linewidth]{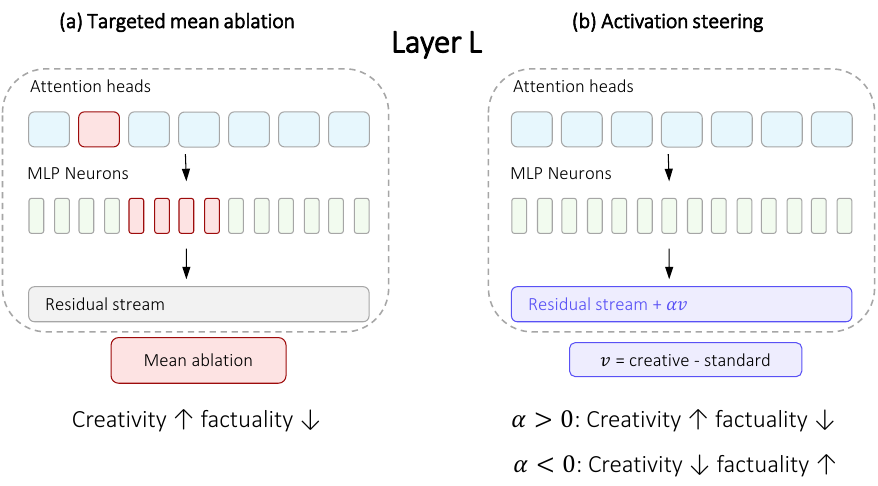}
    \caption{
    Our two intervention methods at layer $L$. (a) Targeted mean ablation replaces a component's contribution with its reference mean $\bar{a}_c$, reducing creativity and improving factuality. (b) Activation steering adds $\alpha v$ to the residual stream at every generated token, with v the mean-pooled difference between creative and standard prompt responses; the sign of $\alpha$ determines the direction of both effects.
    }
    \label{fig:main}
\end{figure}

\section{Related works}
\textbf{Creativity and hallucination in LLMs.}
A growing body of work examines the relationship between creativity and hallucination in LLMs. Early work frames this connection conceptually by arguing that hallucinations should not be understood only as failures of factuality, but also as a possible source of creative novelty \citep{jiang2024survey, chen2025valuable}. Related work further shows that hallucinated outputs can display increased narrativity and semantic coherence \citep{sui2024confabulation}, and that some hallucinations may even be beneficial under creativity-oriented evaluation \citep{sun2025hallucinating_creative}. More recent work studies this relationship directly and empirically. Specifically, empirical evaluations reveal a trade-off between creativity and hallucination across decoding settings \citep{he2025shakespearean_sparks}, while further analysis demonstrates that hallucination-reduction methods can have non-uniform, and sometimes opposite, effects on creativity \citep{banerjee2025less_hallucination_less_creativity}. Furthermore, this relationship has been shown to be nonlinear and task-dependent \citep{yang2025hicbench}. Taken together, these results suggest that creativity and hallucination in LLMs are neither identical nor independent, but partially overlapping behaviors whose interaction depends on how generation is elicited and evaluated.


\textbf{Mechanistic interpretability and activation-space interventions.}
Prior work has used ablations of attention heads and related subcomponents to estimate which parts of a Transformer are functionally important for downstream behavior \citep{michel2019sixteen, wang2022interpretability, yin2025attention}. Other work has shown that factual information can be localized and manipulated through interventions on internal representations \citep{geva2021transformer, meng2022locating, zhong2025react}. Our work builds on this activation-space perspective by using mean ablation \citep{li2024optimal, li2026causal} to test whether components that are important for creativity also play a causal role in hallucination-prone behavior.

\textbf{Activation steering and pooled sequence-level representations.}
Activation steering and representation engineering methods construct control directions from hidden activations and inject them at inference time to modulate model behavior \citep{turner2023steering,panickssery2023steering}. Many such methods use token-specific representations, especially last-token states \citep{panickssery2023steering}. A complementary line of work, however, uses pooled hidden activations when the target property is naturally defined at the level of the full sequence rather than a single token. In these settings, mean pooling is used to obtain a single stable representation per example before constructing contrastive directions or class-conditioned interventions \citep{li2026steering, zhang2025sparse, zou2023representation}. This pattern appears especially in work on global and distributed properties such as emotion and safety, where the goal is to capture sequence-level behavior rather than a token-localized signal \citep{rimsky2024steering, turner2024activation}. Taken together, these studies suggest that pooled representations are a natural choice when the property of interest is expressed across the response as a whole.

\section{Study design}

We study the connection between creativity and hallucinations through two complementary families of interventions, depicted in \Cref{fig:main}. The first ablates model components important for creativity, then tests whether suppressing those same components affects factual performance.
The second uses activation steering to test whether the directions in representation space associated with creative and hallucination-prone behavior are geometrically aligned.
Together, these two approaches provide interventional and geometric evidence supporting a shared mechanism underlying the creativity--factuality trade-off.

\subsection{Models} We run all experiments on Qwen2.5-7B-Instruct \citep{qwen2025qwen25technicalreport} and Llama-3.1-8B-Instruct \citep{grattafiori2024llama3herdmodels}, two instruction-tuned decoder-only Transformer LLMs of comparable scale.
We treat them as representative mid-sized instruction-following models and report results on both throughout.


\subsection{Benchmarks}
\paragraph{Creativity.}
\label{sec:bench_creativity}
We use two story-writing benchmarks that capture complementary aspects of creative generation.
The first is Creative Short Story \citep{ismayilzada2025shortstory}, a cue-word story generation task in which the model is given three words and must write a short story of at most five sentences that incorporates them. The prompt also discourages the most predictable storyline associated with the cue words.
A surprise score is computed on the average semantic distance between consecutive sentences. This score was shown to correlate well with expert judgments of creativity \citep{hou2025creativityprism}.
The second benchmark is CS4 \citep{atmakuru2024cs4}, designed to test whether models can remain creative as prompt specificity increases.
Creativity is evaluated with an LLM-as-a-judge, using GPT-4 \citep{achiam2023gpt}. In addition to these primary story-writing benchmarks, we use two divergent-thinking benchmarks for a secondary generalization evaluation: the Alternative Uses Test (AUT) \citep{guilford1967creativity} and the Divergent Association Task (DAT) \citep{olson2021naming}, following CreativityPrism \citep{hou2025creativityprism}. Further benchmark details are provided in \Cref{app:experimental_details}.

\paragraph{Factuality.}
Throughout this work, we operationalize hallucination as factual inaccuracy on open-domain question-answering benchmarks: a model is said to hallucinate when it produces a fluent but incorrect answer. While this is a narrow definition, and many have suggested alternatives \citep{venkit2024confidently, rawte2023troubling}, it is well-suited to our intervention framework because it yields a precise, measurable signal that can be tracked across ablation and steering conditions.

We evaluate factual performance on two open-domain question-answering benchmarks.
We prompt the model to produce a short answer (See Appendix \ref{sec:eval_factuality}), and report exact match accuracy against reference answers.
The first is TriviaQA \citep{joshi2017triviaqa}, large-scale benchmark of trivia-style question–answer pairs.
The second is Natural Questions (NQ) \citep{kwiatkowski2019natural}, which consists of real queries submitted to a search engine, and is generally more diverse and less uniformly formatted than TriviaQA.

Full benchmark details and evaluation protocols are provided in \Cref{app:experimental_details}. In all intervention experiments, we use the benchmark-specific generation settings from the original papers and report results averaged over five random seeds. As a sanity check, we verify that post-ablation perplexity remains within the baseline distribution, without a substantial deviation relative to its standard deviation, indicating that changes in creativity metrics are not simply driven by degraded generation quality.

\subsection{Targeted component ablation}
\label{sec:background_ablation}
We study component influence through activation-level ablations, a causal intervention framework in which a component's output is replaced with a fixed counterfactual value during inference.
\Cref{fig:main}a demonstrates the method.
In a Transformer block, both attention heads and the MLP produce additive update vectors that are written back into the running token representation via residual addition (up to layer normalization); we refer to this running representation as the residual stream.

Let $a_c(x)$ denote the activation produced by component $c$ on input $x$, representing the additive contribution written to the residual stream at a given layer and token position.
A general ablation intervention takes the form:
\begin{equation}
    a_c(x) \leftarrow \tilde{a}_c
\end{equation}
where $\tilde{a}_c$ is the counterfactual replacement value.
The choice of $\tilde{a}_c$ determines the ablation variant: \emph{zero ablation} sets $\tilde{a}_c = 0$, replacing the component's contribution with zero.
Alternatively, \emph{mean ablation} sets $\tilde{a}_c = \bar{a}_c$, the expected activation over a reference distribution. 

For mean ablation, given a reference distribution $\mathcal{D}$, we compute:
\begin{equation}
    \bar{a}_c = \mathbb{E}_{x' \sim \mathcal{D}}\left[ a_c(x') \right]
\end{equation}

Over all tokens in $\mathcal{D}$.
This approach offers a more conservative estimate of component importance compared to zero ablation, as it maintains the component's typical activation pattern and only removes its context-specific modulation.
Previous work \citep{wang2023ioi} found this strategy to produce more stable results.

\paragraph{Ablation units.}
We treat all attention heads and MLP blocks as candidate components. Each attention head is its own ablation unit. Ablating an entire MLP block proved too coarse, causing a sharp decline in perplexity, so we partition each MLP block into four equal-sized sub-groups and treat each quarter as a separate candidate unit, yielding a finer-grained decomposition without destabilizing the model. E.g., if the hidden dimensionality is 128, we ablate either the first 32 indices, the 2nd, the 3rd or the 4th.


\paragraph{Identifying creativity-critical components.}
For each candidate component $c$, we perform a targeted mean ablation (\Cref{sec:background_ablation}) while holding all other components fixed, and measure the resulting change in creativity score on both benchmarks. The ablation is applied at all token positions, including both the input prompt and generated tokens.

As a sanity check, we verify on both the Creative Short Story and CS4 benchmarks that the perplexity of stories generated under component ablation does not substantially increase relative to stories generated by the original, unablated model. We additionally evaluate model perplexity on WikiText-2 \citep{merity2016pointer}. Full evaluation details and results are provided in Appendix~\ref{app:ppl}.

We rank components by the magnitude of the creativity drop their ablation induces. Because the two creativity benchmarks use different metrics and score scales, we aggregate at the rank level: for each model we combine the two benchmark-specific rankings via average-rank aggregation, equivalent to a Borda-style consensus \citep{dwork2001rank,wang2024survey}.
The top 10 components from this aggregated ranking form the \textit{creativity-critical set}.

\paragraph{Evaluating creativity-critical components on factual QA.}
Having identified the most creativity-critical components, we apply the same mean ablation interventions during inference on TriviaQA and Natural Questions under deterministic decoding, and report the per-component change $\Delta M_{\text{factuality}}(c)$ per benchmark.
To assess whether any observed factuality gains are specific to the creativity-critical components, we compare them against a control distribution obtained by ablating 200 randomly sampled components.

\subsection{Activation steering}
\label{sec:background_steering}

Activation steering modulates model behavior at inference time by adding a learned direction vector to internal representations during the forward pass \citep{klerings2025steering, nadeem2025steering}.
\Cref{fig:main}b demonstrates the method.

We use a contrastive formulation in which steering vectors are derived from two prompt sets associated with opposing behaviors.
Because creativity is expressed across an entire generated response rather than at a single token position, we represent each example by a mean-pooled residual-stream activation.
Let $h_t^{(\ell)}(x)$ denote the residual-stream activation at layer $\ell$ and token position $t$ for a generated output $x$ of length $T_x$.
The pooled representation of $x$ at layer $\ell$ is
\begin{equation}
\bar{h}^{(\ell)}(x)=\frac{1}{T_x}\sum_{t=1}^{T_x} h_t^{(\ell)}(x).
\end{equation}

Given two response sets, positive $\mathcal{R}_+$ and negative $\mathcal{R}_-$, representing model outputs with and without the feature of interest (e.g., creativity).
We construct the steering vector from the difference between their mean pooled activations:
\begin{equation}
v^{(\ell)} = \mathbb{E}_{x\sim\mathcal{R}_+}[\bar{h}^{(\ell)}(x)] - \mathbb{E}_{x\sim\mathcal{R}_-}[\bar{h}^{(\ell)}(x)].
\end{equation}

At inference time, we inject this vector into the residual stream at every generated token position:

\begin{equation}
\tilde{h}_t^{(\ell)} = h_t^{(\ell)} + \alpha v^{(\ell)}.
\end{equation}

\paragraph{Constructing the creativity steering vector.}
We derive a \textit{creativity direction} using the Creative Short Story benchmark.
For each of the 50 cue-word triplets, selected to span a range of semantic distances between words ensuring diversity in cue-word relatedness, we construct two prompt variants differing only in their instruction:
\textit{Creative prompt (original)}, taken directly from the benchmark, Encourages novel, unconventional use of the cue words and explicitly rules out their most obvious interpretation. For example, given petrol, diesel, pump, the prompt specifies that the story should not be about filling a vehicle at a petrol station;
\textit{Standard prompt}, designed by us, asks for a straightforward, coherent story incorporating the same cue words naturally, with no creativity constraint.

We extract residual-stream activations for both variants by mean-pooling over the token positions of each generated response (\Cref{sec:background_steering}), and compute the steering vector as the mean difference between the creative and standard activation pools across all 50 prompt pairs.
Vectors are constructed at every layer; we then select, for each model, the layer whose steering vector produces the largest and most consistent shift in surprise score on a held-out validation set. This yields layer 16 for Qwen and layer 14 for Llama. We verify the stability of the estimated steering directions in
Appendix~\ref{app:steering-convergence}.

\paragraph{Steering toward and away from creativity.}
We apply the creativity steering vector during inference on TriviaQA, sweeping $\alpha$ from negative values (shifting representations away from creativity) to positive values (shifting toward creativity), and measure the effect on exact match (EM) accuracy at each setting.

\paragraph{Geometric alignment between creativity and hallucination directions.}
As a further analysis, we investigate how the \emph{creativity direction} vector is similar to the factuality direction.
To do this, we construct a factuality vector using the same procedure.
We construct the positive and negative response sets by prompting each model with 500 factual questions from TriviaQA, classifying each generated answer as correct or hallucinated (incorrect), using activations extracted at the same layer as the creativity vector.

We then measure the cosine similarity between the hallucination vector $v_{\text{hallucinations}}$ and the creativity vector $v_{\text{creativity}}$.
As a control, we compare this similarity measure against a semantically unrelated direction to verify that any observed alignment is specific to the creativity–hallucination axis rather than an artifact of the representation geometry. 
Concretely, we extract a positive--negative sentiment steering direction from movie-review
prompts from SST-2~\citep{socher2013recursive} and measure its similarity to both the
creativity and hallucination directions.


\paragraph{Generalization evaluation.}
We apply both the creativity and factuality steering vectors to the AUT and DAT benchmarks under negative steering, no steering (baseline) and positive steering. This evaluation tests whether the identified directions generalize beyond creative writing to divergent-thinking tasks.

Full experimental details are provided in \Cref{app:experimental_details}.

\begin{figure}[t]
    \centering
    \begin{subfigure}[t]{0.24\textwidth}
        \centering
        \includegraphics[width=\textwidth]{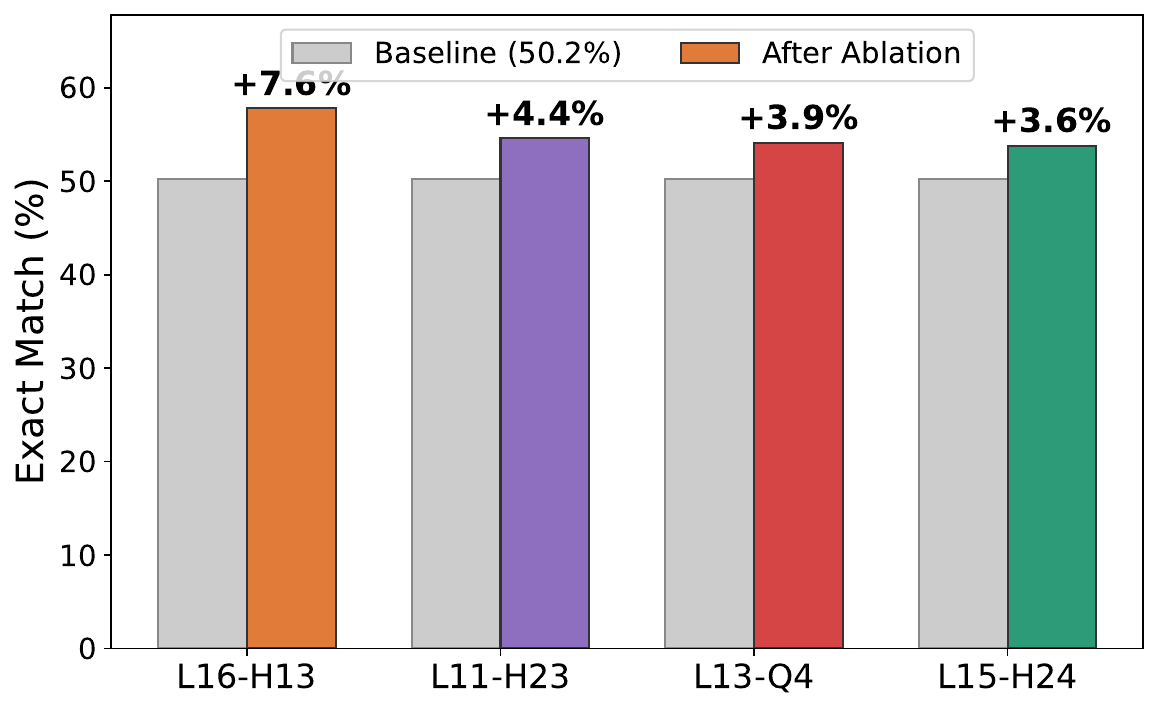}
        \caption{TriviaQA}
    \end{subfigure}
    \hfill
    \begin{subfigure}[t]{0.24\textwidth}
        \centering
        \includegraphics[width=\textwidth]{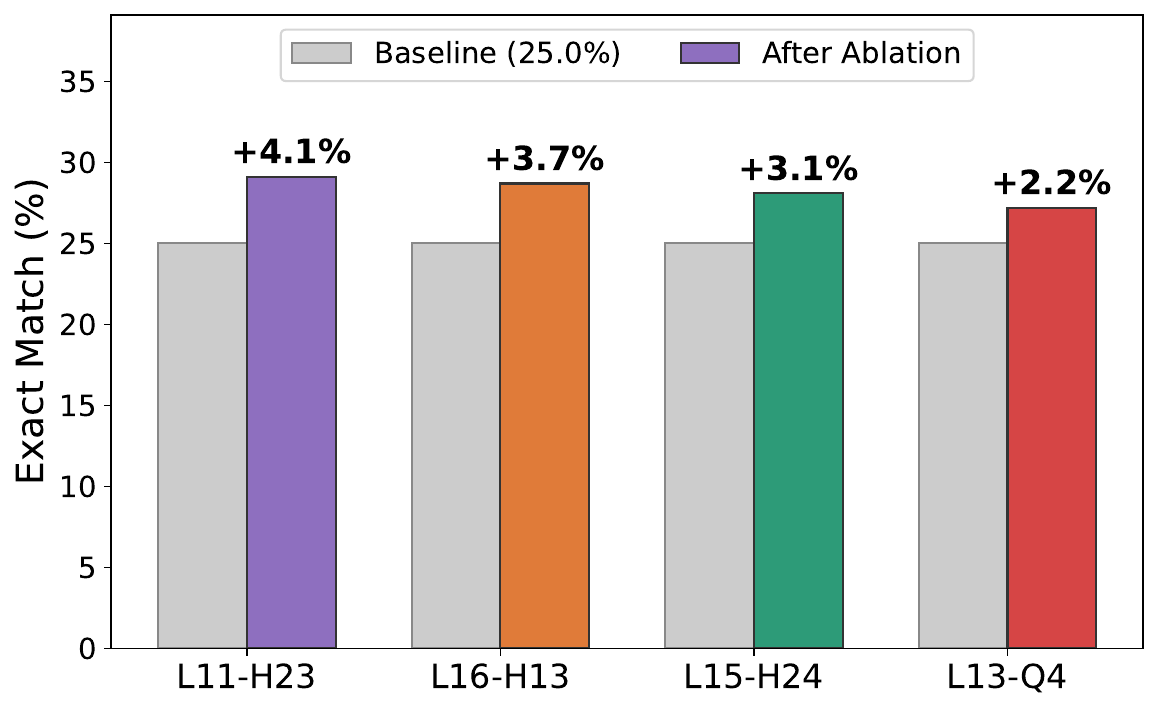}
        \caption{Natural Questions}
    \end{subfigure}
    \hfill
    \begin{subfigure}[t]{0.24\textwidth}
        \centering
        \includegraphics[width=\textwidth]{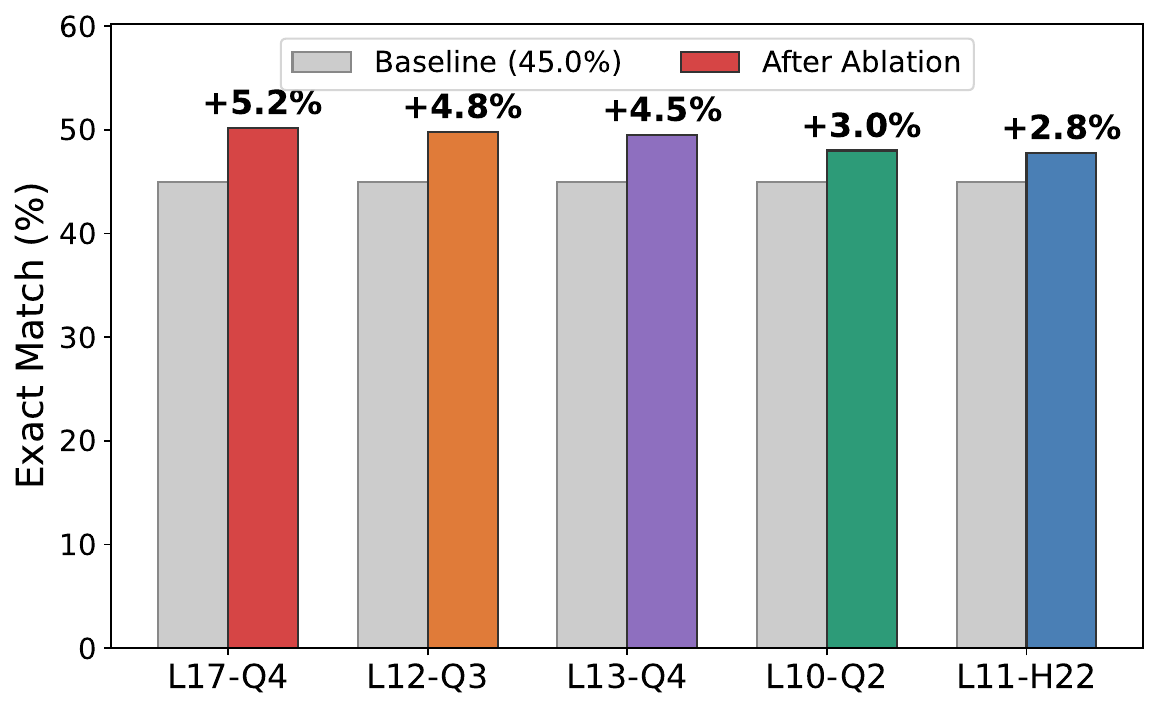}
        \caption{TriviaQA}
    \end{subfigure}
    \hfill
    \begin{subfigure}[t]{0.24\textwidth}
        \centering
        \includegraphics[width=\textwidth]{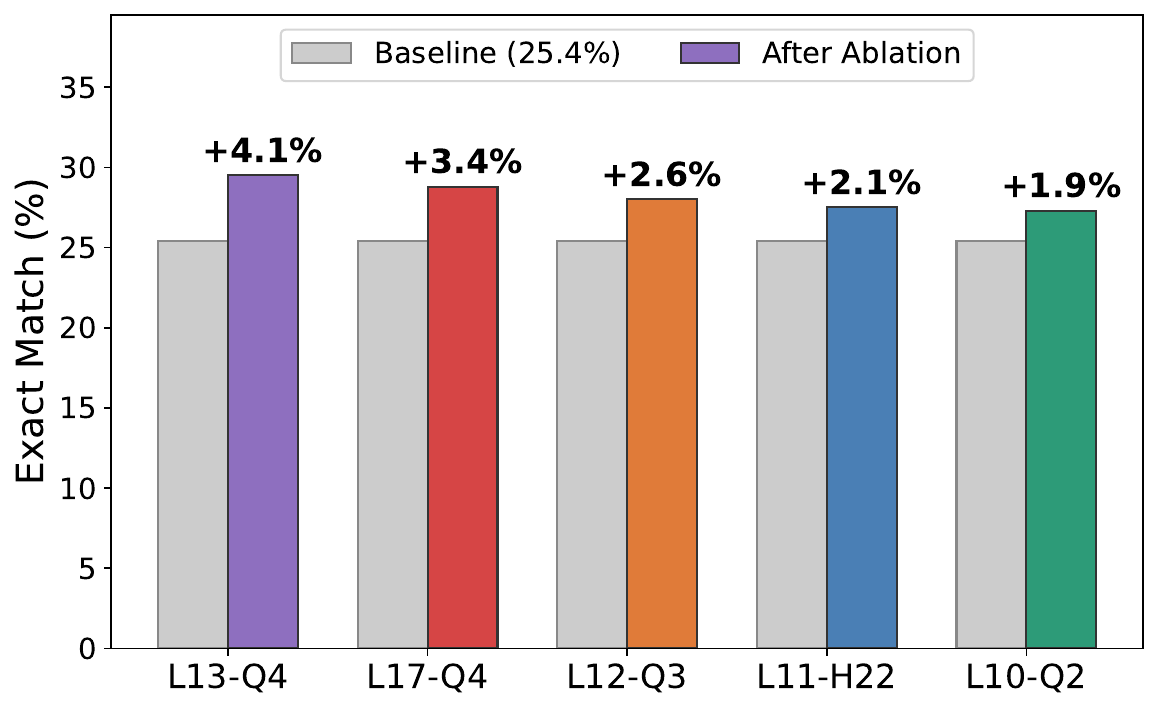}
        \caption{Natural Questions}
    \end{subfigure}

    \vspace{0.5em}
    \makebox[0.48\textwidth][c]{\small Llama-3.1-8B-Instruct}%
    \makebox[0.48\textwidth][c]{\small Qwen2.5-7B-Instruct}
    \caption{
    Factual QA accuracy improves when creativity-critical components are ablated. Each bar mean-ablates one component with all others held fixed; gray bars are the no-ablation baseline and annotations give the absolute gain in exact match accuracy. Results are presented for the components whose gains exceed the random-ablation control.
    }
    \label{fig:hallucination_after_ablation_barplot}
\end{figure}

\section{Results} 
\subsection{Discovering Creativity--Factuality Trade-off Components}



\begin{figure*}[t]
    \centering
    \setlength{\abovecaptionskip}{4pt}
    \setlength{\belowcaptionskip}{0pt}
    \captionsetup[subfigure]{justification=centering,font=small}

    \begin{subfigure}[t]{0.235\linewidth}
        \centering
        \includegraphics[width=\linewidth]{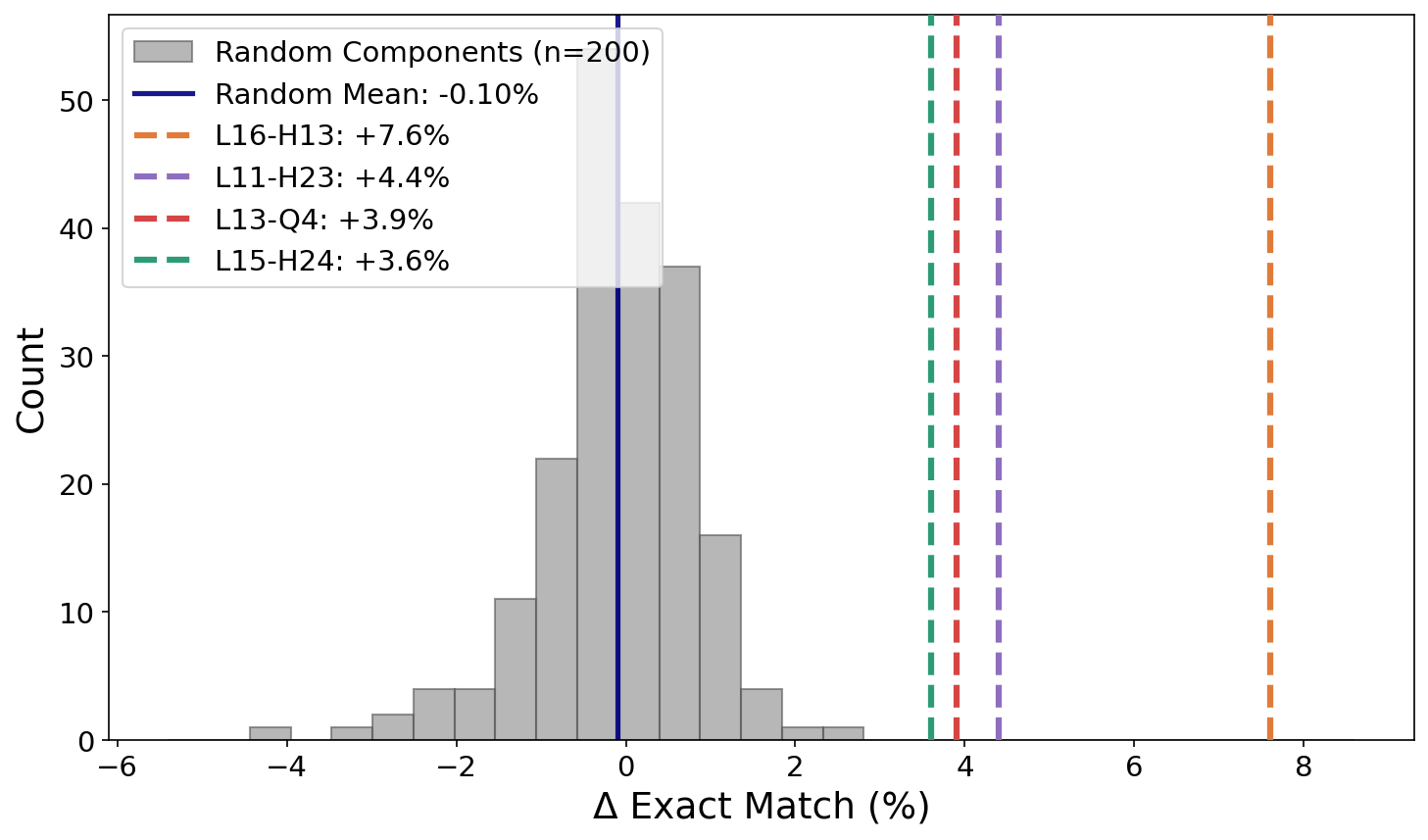}
        \caption{Llama, TriviaQA}
        \label{fig:llama_triviaqa_hist}
    \end{subfigure}
    \hfill
    \begin{subfigure}[t]{0.235\linewidth}
        \centering
        \includegraphics[width=\linewidth]{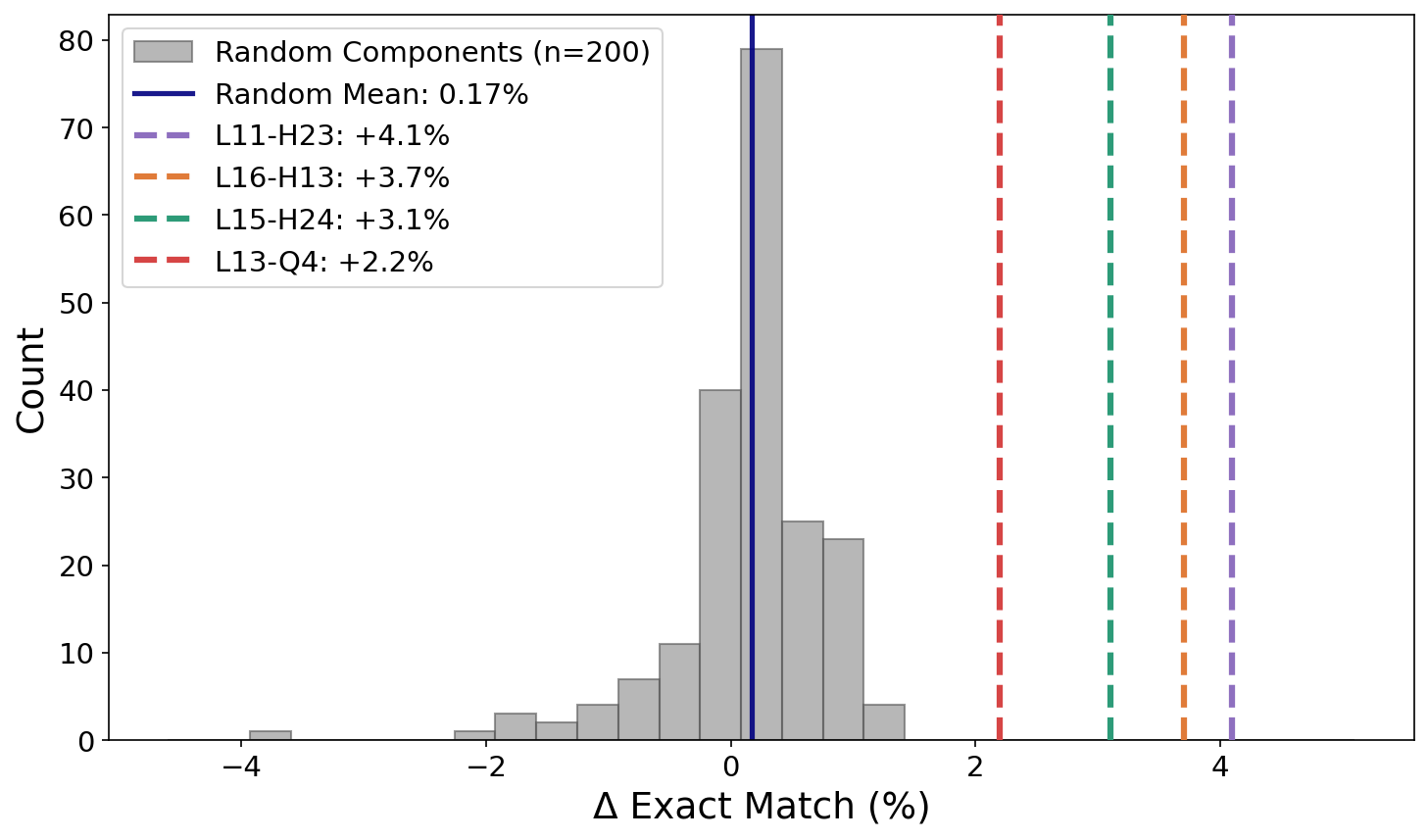}
        \caption{Llama, NQ}
        \label{fig:llama_nq_hist}
    \end{subfigure}
    \hfill
    \begin{subfigure}[t]{0.235\linewidth}
        \centering
        \includegraphics[width=\linewidth]{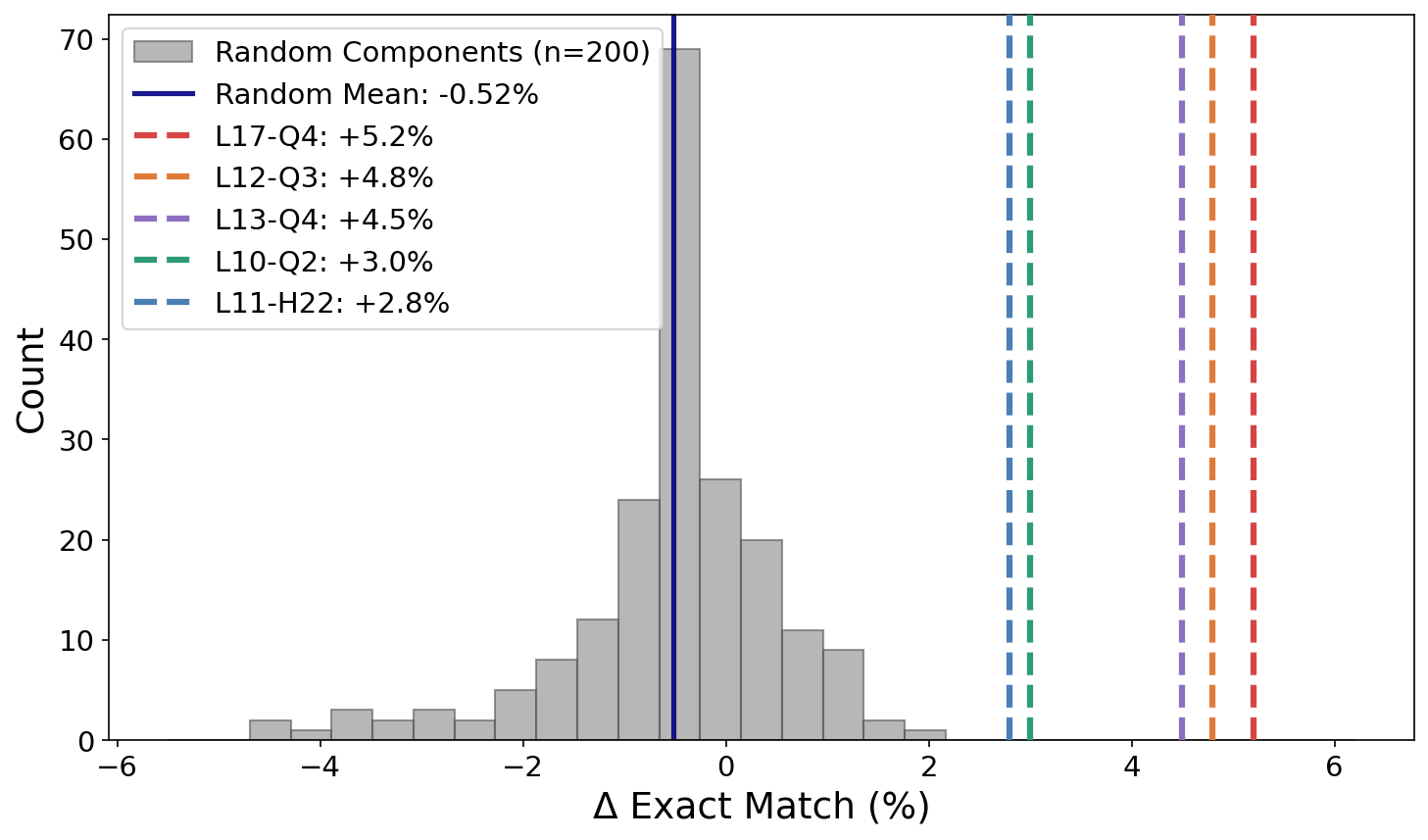}
        \caption{Qwen, TriviaQA}
        \label{fig:qwen_triviaqa_hist}
    \end{subfigure}
    \hfill
    \begin{subfigure}[t]{0.235\linewidth}
        \centering
        \includegraphics[width=\linewidth]{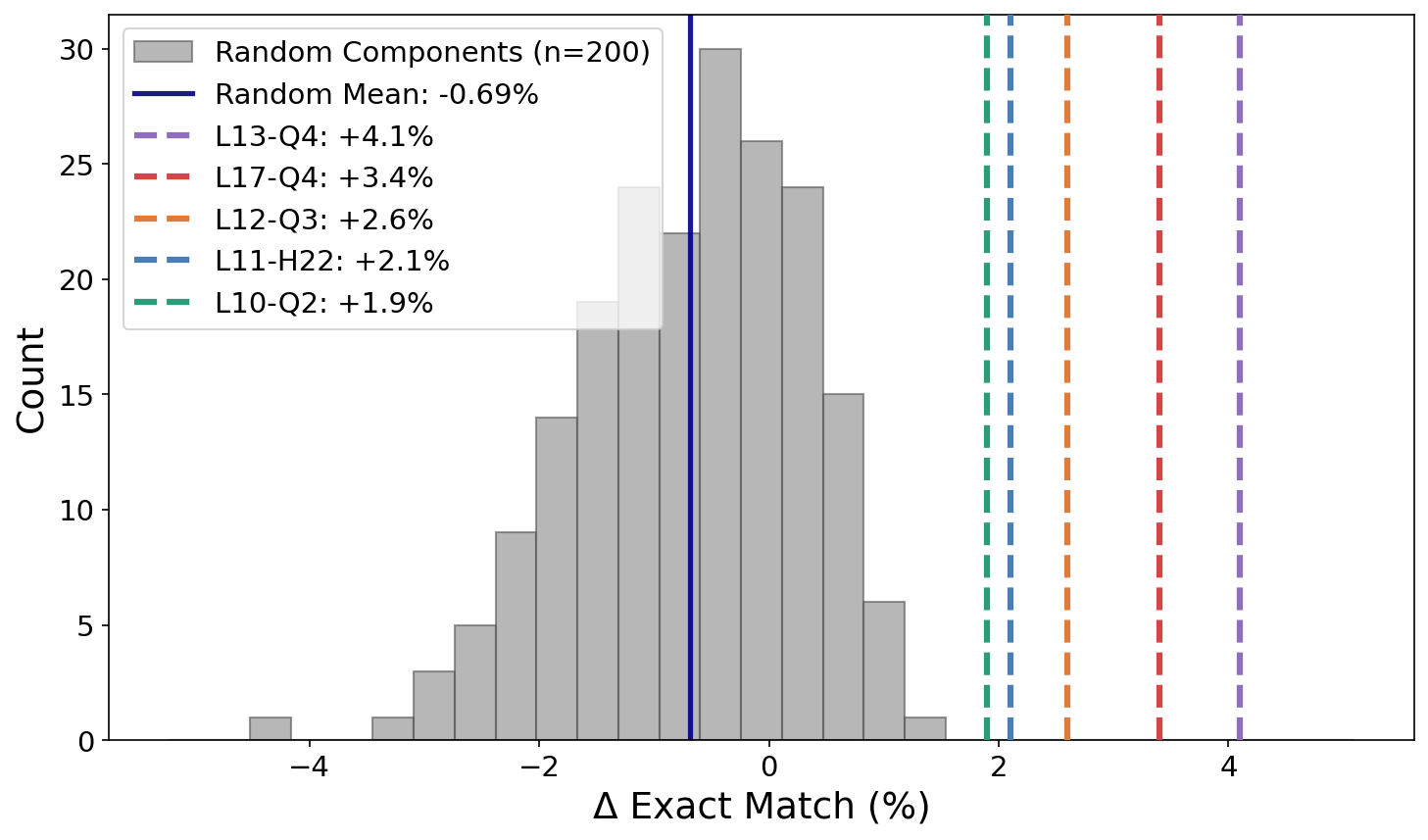}
        \caption{Qwen, NQ}
        \label{fig:qwen_nq_hist}
    \end{subfigure}

    \caption{\textbf{Creativity-critical components vs.\ random components.}
    We show the distribution of changes in EM from ablating 200 random components, compared with selected members of the top-10 creativity-critical set (dashed lines), for Llama-3.1-8B-Instruct and Qwen2.5-7B-Instruct on TriviaQA and Natural Questions. Creativity-critical components produce substantially larger improvements in factual accuracy than random component ablations.}
    \label{fig:fig}
\end{figure*}

\Cref{fig:hallucination_after_ablation_barplot} 
presents the performance of the models on the factuality benchmarks before and after ablating creativity-critical components drawn from the top-10 aggregated creativity ranking. 
We observe a significant increase of factuality across the two benchmarks when ablating these components.
To assess whether the observed improvements could arise from generic ablation rather than the specific functional role of these components, we compare the top-10 creativity components against a control distribution obtained by ablating 200 randomly sampled components. We report here the subset of these components whose factuality gains fall significantly outside the distribution induced by 200 random component ablations.
As shown in Figure \ref{fig:fig}, several components in the top-10 creativity-critical set yield consistent positive shifts in the accuracy on both factuality benchmarks when ablated.
The highlighted creativity-critical components fall well beyond the right tail of this distribution, making a random-ablation explanation unlikely.

As a sanity check, we verified that these factuality gains are not driven by shorter responses. Across both models and factuality benchmarks, response-length differences between baseline and ablated outputs were negligible (Cohen's $d<0.1$ in all conditions; Appendix~\ref{app:length_control}).

This pattern is consistent with a component-level creativity–factuality trade-off: a subset of components that support creative generation also appear to promote hallucination-prone behavior, such that selectively ablating them improves factual reliability.

\subsection{Creativity Steering and Its Effect on Factuality}

\begin{figure}[t]
    \centering
    \begin{subfigure}[t]{0.48\linewidth}
        \centering
        \includegraphics[width=\linewidth]{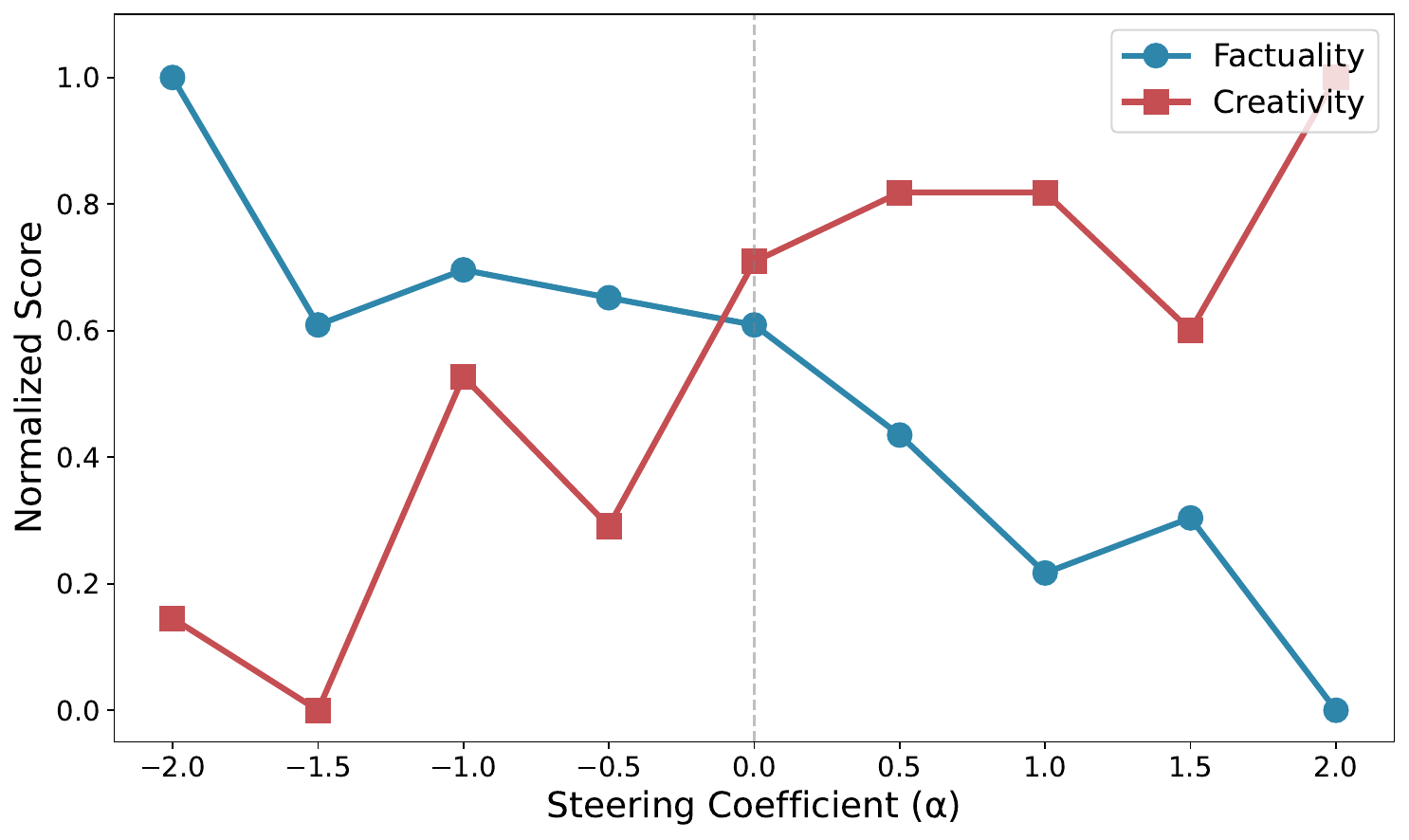}
        \caption{Llama-3.1-8B}
        \label{fig:llama_steering}
    \end{subfigure}
    \hfill
    \begin{subfigure}[t]{0.48\linewidth}
        \centering
        \includegraphics[width=\linewidth]{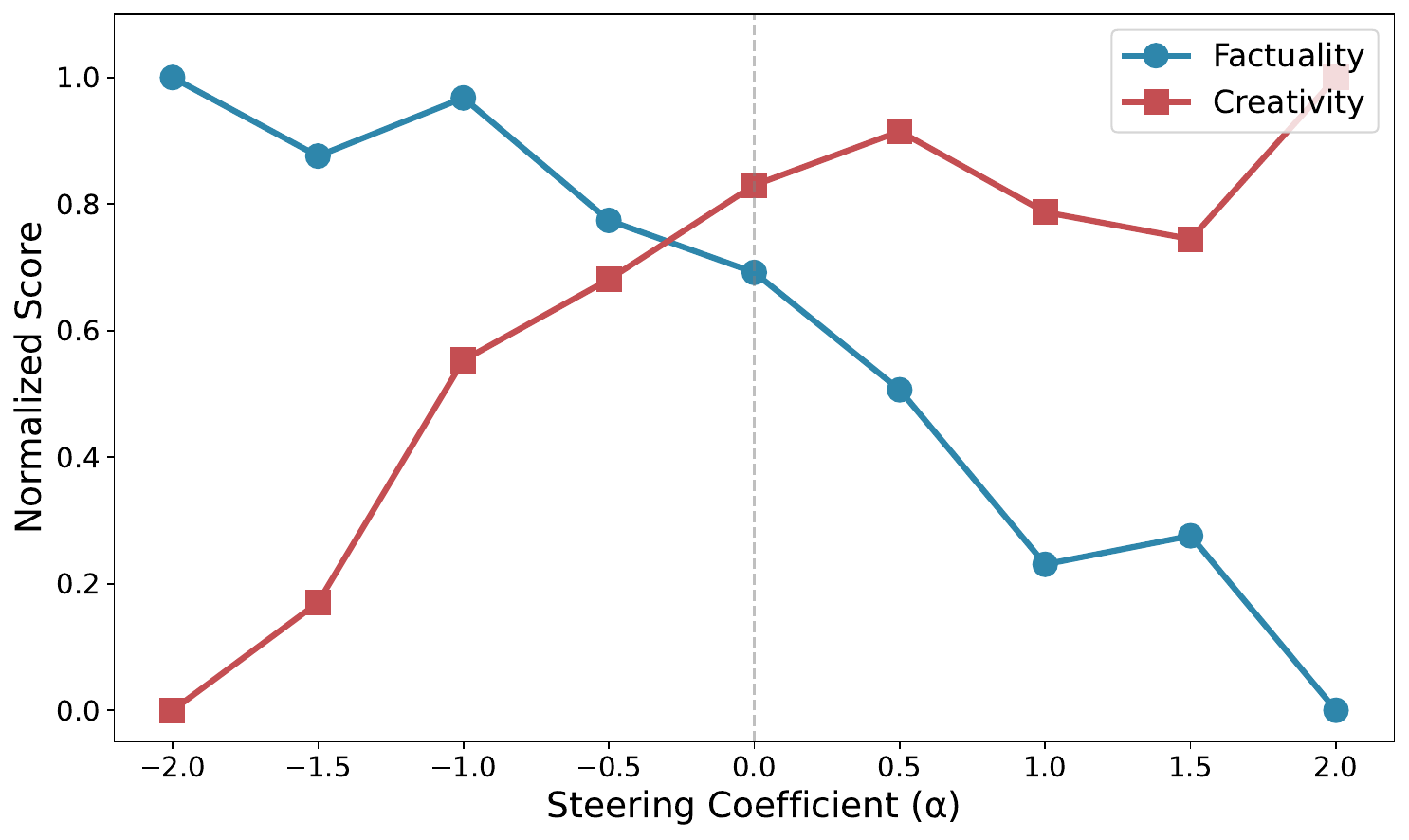}
        \caption{Qwen-2.5-7B}
        \label{fig:qwen_steering}
    \end{subfigure}
    \caption{Effect of a creativity steering vector on two base models. Factuality is measured by Exact Match (EM) on TriviaQA, while creativity is measured by the \emph{surprise} metric on the Creative Short Story benchmark (See \ref{sec:bench_creativity}). Increasing the steering coefficient $\alpha$ consistently raises creativity scores while reducing factual accuracy, revealing a trade-off between creativity and factuality.}
    \label{fig:steering_tradeoff}
\end{figure}

As shown in Figure~\ref{fig:steering_tradeoff}, steering along the creativity direction produces opposing effects on creativity and factuality in both models.
Increasing $\alpha$ raises the surprise scores on the Creative Short Story benchmark, indicating more creative stories, while simultaneously reducing EM accuracy on TriviaQA, consistent with increased hallucination propensity.
Steering in the opposite direction---pushing representations away from creativity---improves factual accuracy at the cost of lower creativity scores.
This bidirectional pattern holds consistently across both Llama and Qwen, drawing a clear trade-off between the two.
Normalized scores in Figure \ref{fig:steering_tradeoff} are obtained by min-max scaling each metric over the range of $\alpha$ tested.
The absolute value difference is presented in Appendix \ref{sec:apndx-steering}.

This result goes beyond the correlational evidence that characterizes prior work: because the same direction vector drives opposing changes in creativity and factuality depending on the sign of $\alpha$, it provides direct evidence that the creativity direction in activation space actively modulates hallucination propensity rather than merely co-varying with it. 
This mirrors the component-level finding and reinforces the view that the two behaviors share a mechanism.

\begin{figure}[t]
    \centering

    \begin{subfigure}[b]{0.49\textwidth}
        \centering
        \includegraphics[width=\linewidth]{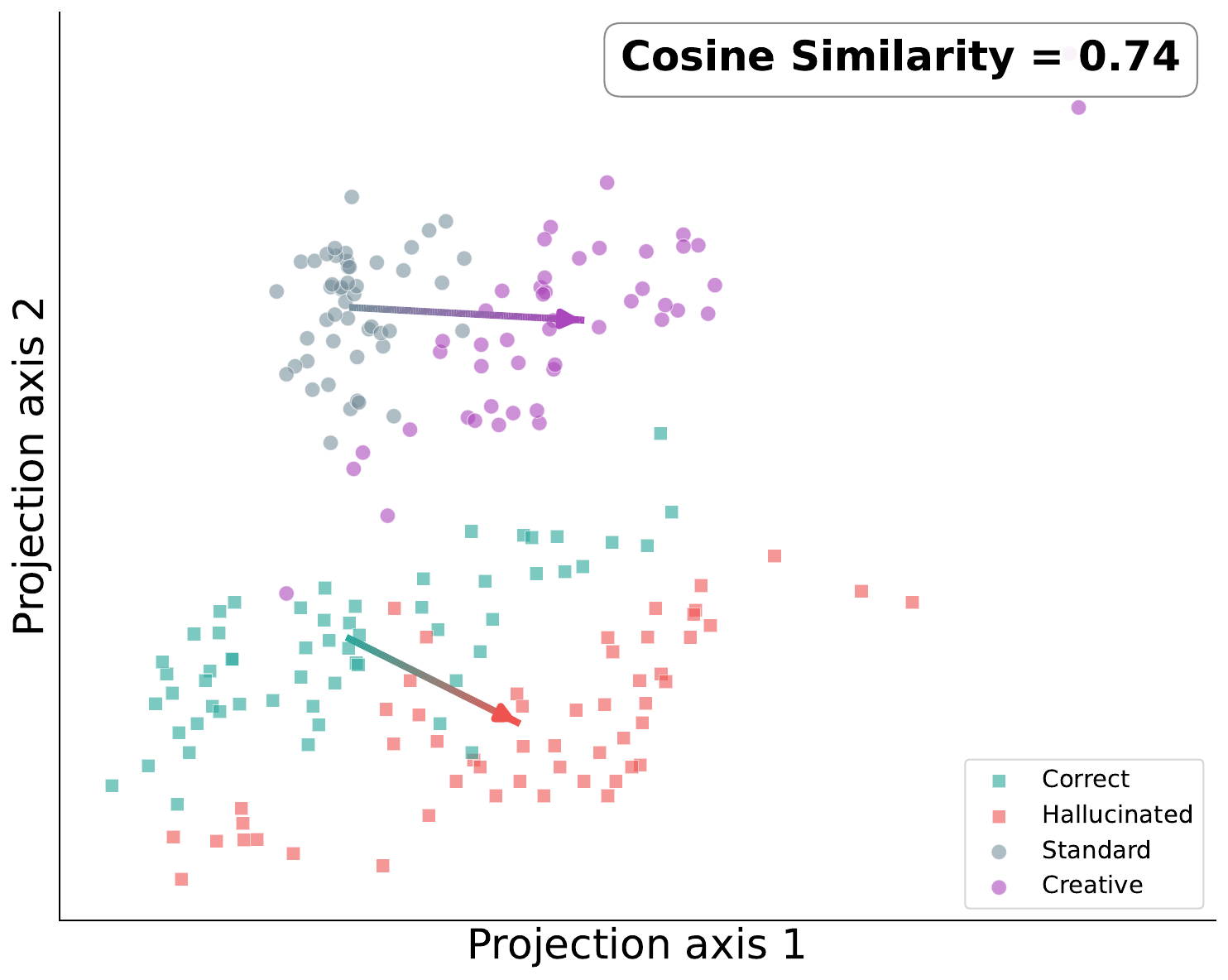}
        \caption{Qwen 2.5-7B-Instruct, Layer 16}
    \end{subfigure}
    \hfill
    \begin{subfigure}[b]{0.49\textwidth}
        \centering
        \includegraphics[width=\linewidth]{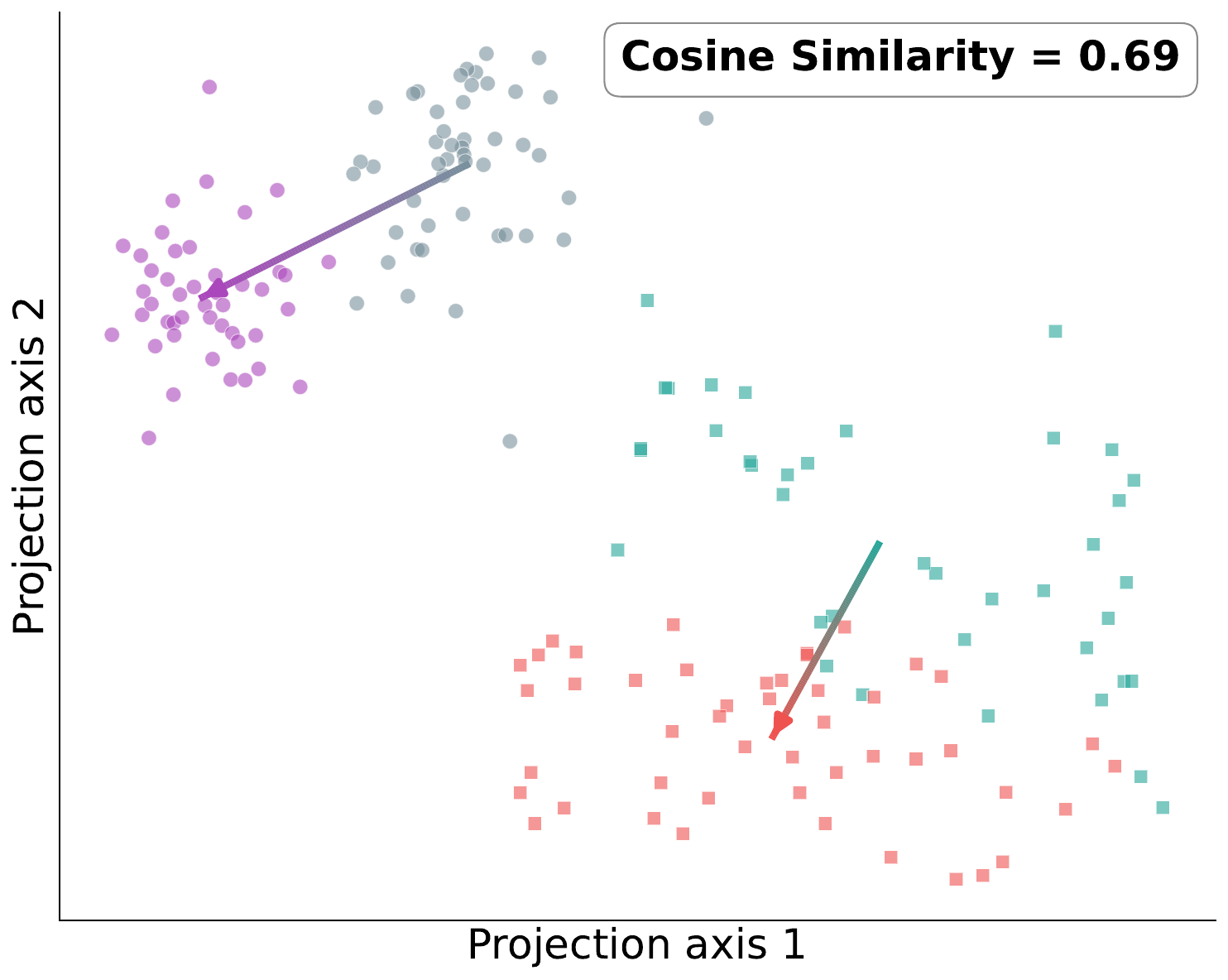}
        \caption{Llama 3.1-8B-Instruct, Layer 14}
    \end{subfigure}

    \caption{\textbf{Creativity and hallucination directions are aligned in activation space.}
    Model activations are projected onto the 2D plane spanned by the creativity and hallucination steering vectors.
    Arrows denote the displacement between the projected class means (Standard $\rightarrow$ Creative, and Correct $\rightarrow$ Hallucinated).
    The reported cosine similarity is computed between the corresponding full-dimensional steering vectors in the original activation space.}

    \label{fig:geometry}
\end{figure}

\paragraph{Geometric Alignment Between Creativity and Hallucination Directions.}
The cosine similarity between the creativity and hallucination steering vectors 
is 0.74 in Qwen and 0.69 in Llama, indicating strong geometric alignment. 
As a control, we compare these vectors to a sentiment baseline direction; 
in contrast to the positive creativity--hallucination alignment, the 
sentiment baseline is near zero or negative in both models (creativity 
vector: $-0.46$ in Qwen, $-0.07$ in Llama; hallucination vector: $-0.39$ 
in Qwen, $-0.13$ in Llama). This suggests the alignment is specific to 
the creativity--hallucination relationship, rather than a generic property 
of arbitrary steering directions. Figure~\ref{fig:geometry} visualizes 
model activations projected onto the plane spanned by the two directions, 
where this alignment is visible geometrically.

\subsection{Generalization Beyond Creative Writing}
\label{sec:generalization}


Across AUT and DAT, steering away from creativity reduces performance in all four settings, while steering away from factuality generally improves it. This pattern holds across both models and suggests that the identified directions extend beyond creative writing to divergent-thinking behavior. Detailed results for each model, benchmark, and steering vector are provided in Appendix~\ref{app:additional_creativity}.

\section{Conclusion and Discussion}

We studied the relationship between creativity and factuality in large language models through three complementary analyses: component-level ablations, activation steering, and representation-space geometry. 
Across all three, we found converging evidence for a shared internal basis underlying the creativity–factuality trade-off.
Components identified as important for creative generation improved factual performance when ablated; steering in the creativity direction increased creativity while reducing factual accuracy; and the creativity and hallucination vectors showed substantial alignment in activation space.
Together, these results indicate that the creativity–factuality tension is not merely a behavioral phenomenon visible in model outputs, but is rooted in the model's internal representations.

These findings advance the mechanistic understanding of the creativity–hallucination connection.
Prior work has established that the two can co-vary under changes in decoding settings or mitigation strategies, but it has remained unclear whether this reflects a genuine causal relationship or a coincidental coupling that could in principle be disentangled.
Our results suggest that creativity and hallucination share overlapping internal computations, making the trade-off a structural feature of how these models are built rather than an artifact of how they are prompted or decoded.
This has a sobering practical implication---reducing hallucinations by targeting creativity---linked components or directions will, at least partially, come at the cost of creative capability, and vice versa.

At the same time, identifying the trade-off precisely is itself a step toward addressing it.
This points toward model designs that more explicitly separate creative and factual modes, avoiding this trade-off.
For instance, through conditional routing, framing in the context, or training objectives that discourage the entanglement we observe here.

Several limitations bound the scope of these conclusions. 
Our experiments cover two models of similar scale and family (Qwen2.5-7B and Llama-3.1-8B); whether the same components and directions play analogous roles in larger models, encoder-decoder architectures, or models trained with different objectives remains to be seen.
Our factuality evaluation also relies on short-answer QA benchmarks, and may not generalize to longer-form or more subtle errors.
Finally, the present analysis focuses primarily on creative writing as the operationalization of creativity; our additional divergent-thinking experiments provide some evidence beyond this setting, but whether similar internal trade-offs arise in other creative domains such as scientific ideation, analogical reasoning, or open-ended problem solving remains an open question.

We identify several potential future directions.
First, refining the components identified here into more localized and interpretable mechanisms, tracing the full circuits through which creativity-linked computations influence factual outputs.
Second, study the geometry of the trade-off in more detail—its layer-wise organization, dependence on prompt and task, and relationship to other behavioral axes in the model.
More broadly, our findings motivate a design-level question: can future architectures be built so that creative and factual computations are more cleanly separated, enabling models that are genuinely capable of both without sacrificing one for the other?
Until then, the creativity–hallucination trade-off should be understood not as a flaw to be patched, but as a structural signature of how current models represent and generate knowledge.

\nocite{langley00}

\bibliography{main}
\bibliographystyle{colm2026_conference}

\newpage
\appendix
\onecolumn

\section{Experimental Details}
\label{app:experimental_details}

\subsection{Benchmarks}
\label{app:benchmarks}
\paragraph{Creative Short Story.}
We use the Creative Short Story benchmark of \citet{ismayilzada2025shortstory}, a cue-word-based story generation task in which the model is given three words and asked to write a creative short story of at most five sentences that incorporates them. The benchmark is designed to elicit controlled but open-ended creative generation: besides requiring the three cue words, the prompt also discourages the most obvious storyline associated with them. Example cue-word sets include \textit{stamp, letter, send}, \textit{petrol, diesel, pump}, and \textit{gloom, payment, exist}.

In our experiments, we focus on the benchmark's \emph{surprise} metric as the primary creativity signal. Intuitively, surprise measures how unexpectedly the story evolves from one sentence to the next. Following the benchmark definition, if a story $S_n$ is segmented into fragments (sentences) $F_1, \dots, F_{|F|}$, then surprise is computed as the average semantic distance between consecutive sentences:
\[
\mathrm{Sur}(S_n)=\frac{2}{|F|-1}\sum_{i=2}^{|F|} \left| D(F_i)-D(F_{i-1}) \right|,
\]
where $D(F_i)$ denotes the semantic representation of fragment $F_i$, and semantic distance is instantiated using embedding-based distance. Operationally, higher surprise corresponds to larger semantic shifts between neighboring sentences, i.e., stories whose development is less predictable as they unfold.

This metric is attractive for our purposes because it provides a fully automatic signal of narrative unexpectedness, rather than relying only on holistic judge ratings. As also noted in later benchmarking work, the Creative Short Story surprise metric captures the novelty of a story's \emph{twist-and-turn trajectory}, rather than simply novelty of its topic or lexical content. We therefore use surprise as our main automatic proxy for creativity on this benchmark. For generation, we use a temperature of 0.7.

\subsection{CS4 Benchmark and Creativity Evaluation}
We use CS4~\citep{atmakuru2024cs4}, a constrained story-writing benchmark designed to test whether models can remain creative as prompt specificity increases. CS4 includes two settings, \textit{instruction-based} and \textit{story-based}, with 50 source instructions in each setting and constraint sets of size 7, 15, 23, 31, and 39, for a total of 500 evaluation prompts. In each instance, the model receives an instruction, a base story, and a set of constraints, and must revise the story to satisfy the constraints while preserving coherence.

Following the original benchmark, evaluation is performed in an LLM-as-a-judge setup. In our experiments, we use the released CS4 code and evaluation pipeline, with GPT-4 as the judge. For generation, we use a temperature of 0.8 and top-$p$ of 0.95.

The released dataset includes real prompt and constraint files for both settings. For example, one instruction-based prompt asks for a story about two characters struggling to shift their priorities and preserve their relationship as they age; its constraint set specifies details such as a small coastal town, careers in science and the arts, distinct health issues, flashbacks to youth, and a meaningful café scene. Another prompt asks for a story about identical twins learning to live separate lives, with constraints including careers in different fields, living far apart, alternating perspectives, and a pivotal mistaken-identity event. These examples illustrate the core design of CS4: evaluating creativity under progressively denser narrative constraints.

\newpage

\paragraph{Divergent-thinking benchmarks.}
For the generalization analysis, we additionally evaluate on two established divergent-thinking benchmarks: the Alternative Uses Test (AUT)~\citep{guilford1967creativity} and the Divergent Association Task (DAT)~\citep{olson2021naming}, following the evaluation setup of CreativityPrism~\citep{hou2025creativityprism}. AUT asks the model to generate unconventional uses for common objects, while DAT asks the model to produce semantically unrelated words and measures creativity based on their semantic distance. These benchmarks provide complementary measures of divergent thinking beyond narrative creativity.

\subsection{Factuality benchmarks}
\label{app:factuality_benchmarks}

We evaluate factuality using two open-domain question answering benchmarks: TriviaQA~\citep{joshi2017triviaqa} and Natural Questions (NQ)~\citep{kwiatkowski2019natural}. In both cases, the model is given a question and asked to generate a short answer, and performance is measured using Exact Match (EM) against the benchmark reference answers.

\paragraph{TriviaQA.}
TriviaQA is a large-scale open-domain QA benchmark built from trivia-style question-answer pairs. Its questions are often knowledge-intensive and entity-centric, requiring retrieval of specific factual information such as names, dates, locations, or titles. Because many items have a relatively well-defined short answer, TriviaQA is a useful testbed for measuring whether an intervention increases or decreases factual reliability under deterministic decoding.

\paragraph{Natural Questions.}
Natural Questions consists of real user questions originally issued to a search engine, and is generally more diverse and less uniformly formatted than trivia-style QA. Compared to TriviaQA, NQ more often includes naturally phrased information-seeking questions and can involve greater variation in answer form. Evaluating on NQ therefore provides a complementary factuality test that is less tied to stylized trivia knowledge.

\subsection{Model-specific notes}
\label{app:model_notes}

We run experiments on two instruction-tuned decoder-only Transformer LLMs: Qwen2.5-7B-Instruct and Llama-3.1-8B-Instruct. Both models use grouped-query attention (GQA) and support long-context inference. Aside from family-specific architectural and tokenizer differences, we treat them as comparable mid-sized instruction-following base models for the purposes of our intervention and steering analyses.

\paragraph{Evaluation protocol.}
\label{sec:eval_factuality}
For both benchmarks, we report Exact Match (EM). A prediction is counted as correct if it exactly matches one of the accepted gold answers after standard normalization. In our intervention experiments, decoding is kept fixed across conditions so that changes in EM can be attributed to the intervention rather than to differences in sampling. For evaluation, we sample 2,000 examples from each factuality benchmark.

To better match the exact-match metric and reduce sensitivity to verbose generations, we use a short-answer prompt, following prior work and model-evaluation guidance on standardizing output format for benchmarking \citep{llama3_herd}. Specifically, we use the following prompt:

Answer the question using a single word or phrase.\\
Question: \texttt{\{question\}}\\
Answer:

\section{Additional Results}

\subsection{Additional Steering results}
\label{sec:apndx-steering}
Table~\ref{tab:steering-results} reports the results of the steering experiment, showing how varying the steering coefficient $\alpha$ affects both factual accuracy (EM) and creativity (Surprise). Negative $\alpha$ values steer toward factuality, while positive values steer toward creativity.
\begin{table*}[t]
\centering
\small

\begin{minipage}[t]{0.46\textwidth}
\centering
\textbf{Llama-3.1-8B (Layer 14)}\\[3pt]
\begin{tabular}{ccc}
\toprule
$\alpha$ & EM (\%) & Surprise \\
\midrule
$-2.0$ & 54.2 & 0.039 \\
$-1.5$ & 50.2 & 0.036 \\
$-1.0$ & 51.1 & 0.047 \\
$-0.5$ & 50.7 & 0.042 \\
$0.0$  & 50.2 & 0.051 \\
$+0.5$ & 48.4 & 0.053 \\
$+1.0$ & 46.2 & 0.053 \\
$+1.5$ & 47.1 & 0.049 \\
$+2.0$ & 44.0 & 0.057 \\
\bottomrule
\end{tabular}
\end{minipage}
\hfill
\begin{minipage}[t]{0.46\textwidth}
\centering
\textbf{Qwen2.5-7B (Layer 16)}\\[3pt]
\begin{tabular}{ccc}
\toprule
$\alpha$ & EM (\%) & Surprise \\
\midrule
$-2.0$ & 50.0 & 0.058 \\
$-1.5$ & 48.0 & 0.062 \\
$-1.0$ & 49.5 & 0.072 \\
$-0.5$ & 46.3 & 0.075 \\
$0.0$  & 45.0 & 0.079 \\
$+0.5$ & 42.0 & 0.081 \\
$+1.0$ & 37.5 & 0.078 \\
$+1.5$ & 38.3 & 0.077 \\
$+2.0$ & 33.8 & 0.083 \\
\bottomrule
\end{tabular}
\end{minipage}

\caption{Steering results for Llama-3.1-8B and Qwen2.5-7B. We report factual accuracy (EM) and creativity (Surprise) across steering coefficients $\alpha$.}
\label{tab:steering-results}
\end{table*}

\subsection{Perplexity Sanity Checks}
\label{app:ppl}

We perform two complementary perplexity-based sanity checks to verify that the observed reductions in creativity under component ablation are not driven by general degradation in generation quality.

First, for each of the Creative Short Story and CS4 benchmarks, we construct a baseline perplexity distribution from stories generated by the original, unablated model. We evaluate generations using the other model as an independent cross-model evaluator: Llama-3.1-8B-Instruct evaluates Qwen2.5-7B-Instruct generations, and Qwen2.5-7B-Instruct evaluates Llama-3.1-8B-Instruct generations. For each ablated component, we compare the perplexity distribution of the generated stories against the corresponding baseline using a one-sided Mann--Whitney $U$ test, with the alternative hypothesis that ablation increases perplexity ($H_1$: $\mathrm{PPL}_{\mathrm{ablated}} > \mathrm{PPL}_{\mathrm{baseline}}$). We do not observe a substantial increase in perplexity under component ablation.

Second, we evaluate model perplexity directly on the WikiText-2 test set using a strided sliding-window approach with context size 2,048 and stride 512. As shown in Table~\ref{tab:wikitext_ppl}, perplexity changes only minimally under ablation for both models.

\begin{table}[t]
\centering
\small
\begin{tabular}{lcc}
\toprule
\textbf{Model} & \textbf{Baseline PPL} & \textbf{Ablated PPL} \\
\midrule
Qwen2.5-7B-Instruct  & 6.501 & $6.512 \pm 0.013$ \\
Llama-3.1-8B-Instruct & 6.395 & $6.414 \pm 0.015$ \\
\bottomrule
\end{tabular}
\caption{Perplexity on the WikiText-2 test set for the original model and under component ablation.}
\label{tab:wikitext_ppl}
\end{table}

Together, these results indicate that the effects observed under component ablation are unlikely to be explained by broad degradation in language modeling quality.

\subsection{Response-Length Control}
\label{app:length_control}

To test whether the factuality improvements under ablation could be explained by shorter responses, we compare response lengths between the no-ablation baseline and the creativity-critical component ablations for both models on TriviaQA and Natural Questions. As shown in Table~\ref{tab:length_control}, response lengths remain nearly unchanged across all conditions, with Cohen's $d<0.1$ for every model--benchmark pair. This suggests that the observed Exact Match improvements are unlikely to be explained by systematic reductions in response length.

\begin{table}[t]
\centering
\small
\setlength{\tabcolsep}{3.5pt}
\begin{tabular}{llrrrrrr}
\toprule
\textbf{Benchmark} & \textbf{Model} &
\shortstack{\textbf{Base.}\\\textbf{Mean}} &
\shortstack{\textbf{Base.}\\\textbf{Std}} &
\shortstack{\textbf{Abl.}\\\textbf{Mean}} &
\shortstack{\textbf{Abl.}\\\textbf{Std}} &
$\boldsymbol{\Delta}$ &
\shortstack{\textbf{Cohen's}\\$\boldsymbol{d}$} \\
\midrule
TriviaQA & Qwen2.5-7B  & 4.36 & 2.26 & 4.29 & 2.15 & $-0.07$ & 0.03 \\
TriviaQA & Llama-3.1-8B & 4.32 & 2.32 & 4.35 & 2.30 & $+0.04$ & 0.04 \\
NQ       & Qwen2.5-7B  & 5.07 & 3.29 & 4.96 & 3.10 & $-0.11$ & 0.03 \\
NQ       & Llama-3.1-8B & 6.13 & 4.47 & 6.29 & 4.70 & $+0.16$ & 0.07 \\
\bottomrule
\end{tabular}
\caption{Mean response length with and without ablation, aggregated over components for each benchmark--model pair. Cohen's $d<0.1$ across all conditions indicates negligible changes in response length.}
\label{tab:length_control}
\end{table}

\subsection{Steering-Vector Convergence}
\label{app:steering-convergence}

To validate the stability of the creativity steering direction with respect to the number of cue-word triplets, we construct steering vectors using subsets of $N\in\{5,10,20,30,40\}$ triplets and compute their cosine similarity to the full vector constructed from all 50 triplets. 

As shown in Table~\ref{tab:steering_convergence}, similarity increases consistently with $N$, from $0.54\pm0.10$ to $0.97\pm0.01$ for Qwen2.5-7B and from $0.48\pm0.10$ to $0.96\pm0.01$ for Llama-3.1-8B. At $N=40$, cosine similarity exceeds $0.96$ for both models, indicating that the estimated direction is highly stable. These results suggest that 50 triplets are sufficient to obtain a robust estimate of the creativity steering direction.

\begin{table}[h]
\centering
\small
\begin{tabular}{ccc}
\toprule
$N$ & \textbf{Qwen2.5-7B} & \textbf{Llama-3.1-8B} \\
\midrule
5  & $0.54 \pm 0.10$ & $0.48 \pm 0.10$ \\
10 & $0.70 \pm 0.05$ & $0.65 \pm 0.05$ \\
20 & $0.87 \pm 0.02$ & $0.82 \pm 0.02$ \\
30 & $0.94 \pm 0.03$ & $0.91 \pm 0.03$ \\
40 & $0.97 \pm 0.01$ & $0.96 \pm 0.01$ \\
\bottomrule
\end{tabular}
\caption{Convergence of the creativity steering direction as a function of the number of cue-word triplets. We report cosine similarity to the steering vector constructed using all 50 triplets.}
\label{tab:steering_convergence}
\end{table}

\newpage

\subsection{Generalization to Additional Creativity Benchmarks}
\label{app:additional_creativity}

To evaluate whether the identified steering directions extend beyond narrative creativity, we consider two established divergent-thinking benchmarks that capture complementary aspects of creative idea generation. The Alternative Uses Test (AUT) \citep{guilford1967creativity} asks for unconventional uses of common objects and evaluates the ability to generate ideas that depart from their typical function. The Divergent Association Task (DAT) \citep{olson2021naming} instead asks for semantically unrelated concepts and measures creativity through the semantic distance between the generated words. 

Thus, whereas our Creative Short Story and CS4 experiments evaluate creativity in narrative generation, AUT and DAT directly target divergent thinking and have been used in several recent evaluations of LLM creativity \citep{hubert2024current, haase2025has, hou2025creativityprism}.

Following CreativityPrism, we evaluate both the creativity and factuality steering vectors on AUT and DAT under negative steering, no steering (baseline), and positive steering. Table~\ref{tab:aut_dat} shows that steering away from creativity decreases performance in all four settings, while steering away from factuality generally improves performance.


The two vectors modulate divergent thinking in opposite directions. For the creativity vector, negative steering lowers performance in all four settings, and positive steering produces the complementary increase in three of four, with the largest change observed on Llama AUT ($19.29 \rightarrow 28.57$). The factuality vector shows the mirror-image pattern: suppressing factuality ($\alpha^-$) raises performance in three of four settings, while reinforcing it ($\alpha^+$) lowers performance throughout, most markedly on Qwen DAT ($68.10 \rightarrow 60.68$). The sole exception in both cases is Llama DAT, where every steered score falls within $0.5$ points of the $74.65$ baseline. Overall, these results show that steering toward creativity and steering away from factuality have convergent effects on divergent-thinking behavior, while the opposite interventions suppress it, extending the observed relationship beyond creative writing.

\begin{table}[t]
\centering
\small
\setlength{\tabcolsep}{5pt}
\begin{tabular}{llcrrr}
\toprule
\textbf{Vector} & \textbf{Model} & \textbf{Benchmark} &
$\boldsymbol{\alpha^-}$ & \textbf{Base.} & $\boldsymbol{\alpha^+}$ \\
\midrule
Creativity & Qwen  & AUT & 9.81  & 10.08 & 10.51 \\
Creativity & Qwen  & DAT & 65.45 & 68.10 & 73.67 \\
Creativity & Llama & AUT & 16.24 & 19.29 & 28.57 \\
Creativity & Llama & DAT & 70.60 & 74.65 & 74.36 \\
\midrule
Factuality & Qwen  & AUT & 10.97 & 10.08 & 9.95  \\
Factuality & Qwen  & DAT & 70.61 & 68.10 & 60.68 \\
Factuality & Llama & AUT & 22.38 & 19.29 & 18.33 \\
Factuality & Llama & DAT & 74.53 & 74.65 & 74.15 \\
\bottomrule
\end{tabular}
\caption{Generalization of the creativity and factuality steering directions to the Alternative Uses Test (AUT) and Divergent Association Task (DAT). We report performance under negative steering ($\alpha^-$), no steering (baseline), and positive steering ($\alpha^+$).}
\label{tab:aut_dat}
\end{table}

\newpage

\subsection{Qualitative Examples of Factual Corrections}
\label{sec:qualitative-examples}



\paragraph{Llama 3.1-8B-Instruct}

\subsubsection*{Layer 11, Attention Head 23 (\texttt{layer11\_h23})}

\begin{longtable}{|c|l|p{3cm}|p{2.8cm}|p{2.8cm}|p{1.9cm}|}
\hline
\# & DS & Question & Baseline (Wrong) & After Ablation (Correct) & Ground Truth \\
\hline
1 & TriviaQA & Born in 1861, which opera singer's real name was Helen Mitchell? & Emma Calvé, a French soprano & Nellie Melba & Nellie Melba \\
\hline
2 & TriviaQA & What is a nervous disorder whereby a person suddenly falls asleep? & Cataplexy & Narcolepsy & Narcolepsy \\
\hline
3 & TriviaQA & Who plays the role of Bubble in Absolutely Fabulous? & Kate Moss & Jane Horrocks & Jane Horrocks \\
\hline
4 & TriviaQA & Which Roxy Music hit single features Robert E Lee? & More Than This & Virginia Plain & Virginia Plain \\
\hline
5 & TriviaQA & In Star Wars, what is the name of the slug-like alien gangster? & Greedo & Jabba the Hutt & Jabba the Hutt \\
\hline
\end{longtable}

\subsubsection*{Layer 16, Attention Head 13 (\texttt{layer16\_h13})}

\begin{longtable}{|c|l|p{3cm}|p{2.8cm}|p{2.8cm}|p{1.9cm}|}
\hline
\# & DS & Question & Baseline (Wrong) & After Ablation (Correct) & Ground Truth \\
\hline
1 & NQ & who co-created fred ott's sneeze & George Albert Smith & William Kennedy Dickson & W.K. Dickson \\
\hline
2 & TriviaQA & The largest volcano in our solar system is on which planet? & Venus & Mars (Olympus Mons) & Mars \\
\hline
3 & TriviaQA & Which canal joins the North Sea at Brunsbüttel? & The Elbe River & Nord-Ostsee-Kanal (Kiel Canal) & Kiel Canal \\
\hline
4 & TriviaQA & When was the first road speed limit set in the UK? & 1896 & 1861 & 1861 \\
\hline
5 & TriviaQA & What is 1009 when expressed as Roman numerals? & CIX & MIX & MIX \\
\hline
\end{longtable}

\subsubsection*{Layer 13, MLP Quarter 4 (\texttt{layer13\_q4})}

\begin{longtable}{|c|l|p{3cm}|p{2.8cm}|p{2.8cm}|p{1.9cm}|}
\hline
\# & DS & Question & Baseline (Wrong) & After Ablation (Correct) & Ground Truth \\
\hline
1 & NQ & when does a wrinkle in time come out in canada & A 2018 film by Ava DuVernay & March 9, 2018 & March 9, 2018 \\
\hline
2 & NQ & how many house of representatives in nigeria & Nigeria has a unicameral legislature & 360 members & 360 \\
\hline
3 & TriviaQA & Which Oscar-nominated film had You Sexy Thing as theme? & Mamma Mia! & The Full Monty & The Full Monty \\
\hline
4 & TriviaQA & Men Against the Sea and Pitcairn's Island were sequels to? & Typee by Melville & Mutiny on the Bounty & Mutiny on the Bounty \\
\hline
5 & TriviaQA & Which American artist painted `Cape Cod Evening'? & Winslow Homer & Edward Hopper & Edward Hopper \\
\hline
\end{longtable}

\subsubsection*{Layer 15, Attention Head 24 (\texttt{layer15\_h24})}

\begin{longtable}{|c|l|p{3cm}|p{2.8cm}|p{2.8cm}|p{1.9cm}|}
\hline
\# & DS & Question & Baseline (Wrong) & After Ablation (Correct) & Ground Truth \\
\hline
1 & NQ & who did kenny rogers sing mary did you know with & Michael English & Wynonna Judd & Wynonna Judd \\
\hline
2 & TriviaQA & What was Joan Jett's backing group? & The Runaways & The Blackhearts & The Blackhearts \\
\hline
3 & TriviaQA & The largest volcano in our solar system is on which planet? & Venus & Mars (Olympus Mons) & Mars \\
\hline
4 & TriviaQA & The Great American Chocolate Bar is better known as? & Snickers & Hershey's & Hershey's \\
\hline
5 & TriviaQA & Who won women's Skeleton Bob gold at 2010 Winter Olympics? & Amy Peterson & Amy Williams & Amy Williams \\
\hline
\end{longtable}

\paragraph{Qwen 2.5-7B-Instruct}

\subsubsection*{Layer 17, MLP Quarter 4 (\texttt{layer17\_q4})}

\begin{longtable}{|c|l|p{3cm}|p{2.8cm}|p{2.8cm}|p{1.9cm}|}
\hline
\# & DS & Question & Baseline (Wrong) & After Ablation (Correct) & Ground Truth \\
\hline
1 & TriviaQA & What Latin phrase, which translates to English as "for this", is taken to mean for the particular purpose? & Propter hoc & Ad hoc & Ad hoc \\
\hline
2 & TriviaQA & What hill does a bill sit on before it becomes a law? & A bill sits on the Senate and House floors... & Capitol Hill & Capitol Hill \\
\hline
3 & TriviaQA & How many champagne bottles are there in a Methuselah? & 6 champagne bottles & 8 champagne bottles & 8 \\
\hline
4 & TriviaQA & Approximately what percentage of an incandescent light bulb's energy is wasted as heat? & 80\% & 90\% & 90\% \\
\hline
5 & NQ & Host of the Late Show who was once a correspondent for The Daily Show & Trevor Noah & Stephen Colbert & Stephen Colbert \\
\hline
\end{longtable}

\subsubsection*{Layer 12, MLP Quarter 3 (\texttt{layer12\_q3})}

\begin{longtable}{|c|l|p{3cm}|p{2.8cm}|p{2.8cm}|p{1.9cm}|}
\hline
\# & DS & Question & Baseline (Wrong) & After Ablation (Correct) & Ground Truth \\
\hline
1 & TriviaQA & What was the name of the character played by Ralph Fiennes in \textit{Schindler's List}? & Oskar Schindler & Amon Göth & Amon Göth \\
\hline
2 & TriviaQA & In fashion, Hobble, Pencil and Dirndl are all types of what? & clothing & skirt & skirt \\
\hline
3 & TriviaQA & In which country was the 2006 FIFA World Cup held? & Italy & Germany & Germany \\
\hline
4 & TriviaQA & Can you name the singer of the title track in \textit{GoldenEye} (1995)? & Chris De Burgh & Tina Turner & Tina Turner \\
\hline
5 & NQ & Who played Mike Stivic on \textit{All in the Family}? & Gary Burghoff & Rob Reiner & Rob Reiner \\
\hline
\end{longtable}

\subsubsection*{Layer 13, MLP Quarter 4 (\texttt{layer13\_q4})}

\begin{longtable}{|c|l|p{3cm}|p{2.8cm}|p{2.8cm}|p{1.9cm}|}
\hline
\# & DS & Question & Baseline (Wrong) & After Ablation (Correct) & Ground Truth \\
\hline
1 & TriviaQA & John Chapter 11 Verse 35 is the shortest verse in the Bible. How many words make up the verse? & One word & Two words & 2 \\
\hline
2 & TriviaQA & What is added to white wine to make a Kir? & Blackcurrant juice & Crème de cassis & Cassis \\
\hline
3 & TriviaQA & In which country was the 2006 FIFA World Cup held? & Italy & Germany & Germany \\
\hline
4 & TriviaQA & In fashion, Hobble, Pencil and Dirndl are all types of what? & clothing & skirts & skirt \\
\hline
5 & NQ & What is the average salary for a US congressman? & \$17,000 & \$174,000 & \$174,000 \\
\hline
\end{longtable}

\subsubsection*{Layer 10, MLP Quarter 2 (\texttt{layer10\_q2})}

\begin{longtable}{|c|l|p{3cm}|p{2.8cm}|p{2.8cm}|p{1.9cm}|}
\hline
\# & DS & Question & Baseline (Wrong) & After Ablation (Correct) & Ground Truth \\
\hline
1 & TriviaQA & In darts, what is the highest possible checkout with three darts, finishing with a bull? & 180 & 170 & 170 \\
\hline
2 & TriviaQA & In which country was the 2006 FIFA World Cup held? & Italy & Germany & Germany \\
\hline
3 & TriviaQA & Based on market cap, who was declared the top technology company? & Tesla & Apple & Apple \\
\hline
4 & TriviaQA & Which racehorse, unbeaten in 14 races, won 9 consecutive Group 1 races? & Cape Cross & Frankel & Frankel \\
\hline
5 & NQ & Who played Marius in the movie \textit{Les Misérables}? & Ewan McGregor & Eddie Redmayne & Eddie Redmayne \\
\hline
\end{longtable}

\subsubsection*{Layer 11, Attention Head 22 (\texttt{layer11\_h22})}

\begin{longtable}{|c|l|p{3cm}|p{2.8cm}|p{2.8cm}|p{1.9cm}|}
\hline
\# & DS & Question & Baseline (Wrong) & After Ablation (Correct) & Ground Truth \\
\hline
1 & TriviaQA & In which decade of the 17th Century was the New Model Army created? & 1645 & 1640s & 1640s \\
\hline
2 & TriviaQA & What was the first name of the father of the Brontë sisters? & William & Patrick & Patrick \\
\hline
3 & TriviaQA & Born ``Lucille LeSueur'', by what name is this actress better known? & Bette Davis & Joan Crawford & Joan Crawford \\
\hline
4 & TriviaQA & In 1965, who became the first American astronaut to walk in space? & Eugene Cernan & Edward White & Edward White \\
\hline
5 & TriviaQA & In which year was the Royal British Legion founded? & 1925 & 1921 & 1921 \\
\hline
\end{longtable}

\end{document}